\documentclass[runningheads]{llncs}

\usepackage{algorithm}
\usepackage{algorithmic}
\usepackage{amsmath}
\usepackage{amssymb}
\usepackage{array}
\usepackage{arydshln}

\usepackage{color}
\usepackage{comment}
\usepackage{dsfont}
\usepackage{epsfig}
\usepackage{graphicx}
\usepackage{mathrsfs}
\usepackage{caption}
\usepackage{subcaption}
\usepackage{verbatim}
\usepackage{overpic}

\newcommand{\R}{\mathbb{R}}
\let\bs=\boldsymbol

\def \diag {\mathrm{diag}}
\def \det {\mathrm{det}}

\def \saliency {\textup{\saliency}}

\def \path {\mathit{path}}

\def \label {\mathit{label}}

\def \view {\textup{view}}

\def \labeled {\textup{labeled}}

\def \aligned {\textup{align}}

\let\set=\mathcal

\global\long\def\diag{\mathrm{diag}}

\global\long\def\R{\mathbb{R}}

\usepackage[pagebackref=true,breaklinks=true,letterpaper=true,colorlinks,bookmarks=false]{hyperref}

\usepackage{graphicx}
\usepackage{amsmath,amssymb} 
\usepackage{color}
\usepackage[width=122mm,left=12mm,paperwidth=146mm,height=193mm,top=12mm,paperheight=217mm]{geometry}

\newtheorem{prop}{\textbf{Proposition}}

\numberwithin{theorem}{section}
\numberwithin{lem}{section}

\begin{document}

\newcommand\linjie[1]{\textcolor{red}{Linjie: #1}}
\newcommand\qixing[1]{\textcolor{green}{Qixing: #1}}
\newcommand\Xingyi[1]{\textcolor{blue}{Xingyi: #1}}
\newcommand\Angela[1]{\textcolor{red}{Angela: #1}}
\newcommand\hide[1]{}
\newcommand\TODO[1]{\textcolor{red}{TODO: #1}}
\let\bs=\mathbf

\pagestyle{headings}
\mainmatter

\title{Unsupervised Domain Adaptation for 3D Keypoint Estimation via View Consistency} 

\titlerunning{3D Keypoints via View Consistency}

\authorrunning{Zhou X., Karpur A., Gan, C., Luo L., Huang Q.}

\author{Xingyi Zhou\inst{1} \and Arjun Karpur\inst{1}
\and Chuang Gan\inst{2} \and Linjie Luo\inst{3} \and Qixing Huang\inst{1}}


\institute{The University of Texas at Austin \and
    MIT-IBM Watson AI Lab \and
	Snap Inc. \\
	\email{ \{zhouxy, akarpur, huangqx\}@cs.utexas.edu, \\ ganchuang1990@gmail.com, linjie.luo@snap.com}
}

\maketitle

\begin{abstract}
    In this paper, we introduce a novel unsupervised domain adaptation technique for the task of 3D keypoint prediction from a single depth scan or image. Our key idea is to utilize the fact that predictions from different views of the same or similar objects should be consistent with each other. Such view consistency can provide effective regularization for keypoint prediction on unlabeled instances. In addition, we introduce a geometric alignment term to regularize predictions in the target domain. The resulting loss function can be effectively optimized via alternating minimization. We demonstrate the effectiveness of our approach on real datasets and present experimental results showing that our approach is superior to state-of-the-art general-purpose domain adaptation techniques. 

\end{abstract}

\section{Introduction}
\label{Section:Introduction}

A new era has arrived with the proliferation of depth-equipped sensors in all kinds of form factors, ranging from wearables and mobile phones to on-vehicle scanners. This ever-increasing amount of depth scans is a valuable resource that remains largely untapped, however, due to a lack of techniques capable of efficiently processing, representing, and understanding them. 

3D keypoints, which can be inferred from depth scans, are a compact yet semantically rich representation of 3D objects that have proven effective for many tasks, including reconstruction~\cite{guptaCVPR15a}, object segmentation and recognition~\cite{li2015database}, as well as pose estimation~\cite{DBLP:journals/corr/TulsianiM14}. Despite the wide availability of depth scans of various object categories~\cite{Choi2016}, there is a lack of corresponding 3D keypoint annotations, which are necessary to train reliable keypoint predictors in a supervised approach. This is partially due to the fact that depth scans are inherently partial views of the underlying objects, making it difficult to annotate the object parts occluded from view. One could automate the annotation process by leveraging the ``fused'' models created using the depth scans, but most depth-fusion methods are susceptible to scanning noise and cascading errors when depth scans are captured at scale~\cite{Choi2016}.

\begin{figure}[t]
\centering
\setlength{\tabcolsep}{0pt}
\resizebox{0.8\columnwidth}{!}{%
\begin{tabular}{cccccccc}
    \includegraphics[trim=45 0 45 0,clip,width=0.11\linewidth]{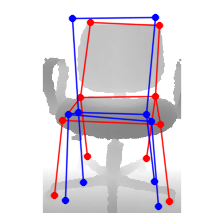} &
    \includegraphics[trim=45 0 45 0,clip,width=0.11\linewidth]{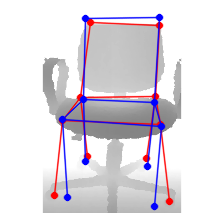} &
    
    \includegraphics[trim=22 0 22 0,clip,width=0.15\linewidth]{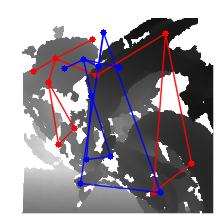} &
    \includegraphics[trim=22 0 22 0,clip,width=0.15\linewidth]{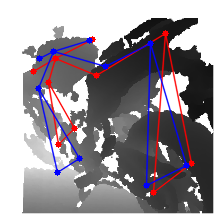} &
\end{tabular}
}
\caption{Our approach improves 3D keypoint prediction results from single depth scans of the Redwood dataset~\protect\cite{Choi2016}. For each pair: \textbf{(Left)} without domain adaptation, the pre-trained keypoint predictor from simulated examples failed to predict accurate 3D keypoints (blue). \textbf{(Right)} 3D keypoint predictions (blue) after domain adaptation are significantly improved. Note that the ground-truth keypoints are shown in red for comparison.}
\label{Fig:Best:Result}
\end{figure}

In this paper, our goal is to predict 3D keypoints of an underlying object from a single raw depth scan. To train a reliable 3D keypoint predictor, we generate a large dataset of simulated depth scans using large-scale 3D model repositories such as ShapeNet~\cite{DBLP:journals/corr/ChangFGHHLSSSSX15} and ModelNet~\cite{conf/cvpr/WuSKYZTX15}. The 3D keypoint annotations on the 3D models from these repositories can naturally carry over to the simulated depth scans for effective supervised training. A large gap exists, however, between the simulated and real depth scan domains. Particularly, 3D models from repositories are generally designed with interactive tools, inevitably resulting in inaccurate geometries with varying scales. Furthermore, the real depth scans contain noticeable measurement noise and background objects, and the class distributions of 3D models from the repositories and those from real depth scans may be quite different.

To close the gap between the source domain of simulated depth scans and the target domain of real depth scans, we introduce a novel approach for unsupervised domain adaptation of 3D keypoint prediction. Our approach is motivated by the special spatial properties of the 3D keypoints and the relationship between the keypoint distributions of the source and target domains.

First, keypoint predictions from different views of the same 3D model should be consistent with each other up to a pose transformation. This allows us to formulate a \emph{view-consistency} regularization to propagate a good prediction, e.g. from a well-posed view where the prediction is more accurately adapted, to a challenging view with less accurate adaptation. To this end, we introduce a latent keypoint configuration to fuse the keypoint predictions from different views of the same object. Additionally, we introduce a pose-invariant metric to compare the keypoint predictions, which allows us to leverage depth scans without camera pose calibration for training.

Second, despite the distinctive differences between the source and target domains, their 3D keypoint distributions are highly correlated. However, naively aligning the 3D keypoint distributions between the two domains is sub-optimal since the occurrences of the same type of objects differ. To address this challenge, we propose a \emph{geometric alignment} regularization that is insensitive to varying densities of the objects in order to align the keypoint distributions of the two domains. We make use of the target domain's latent keypoint configurations from view consistency regularization to compute the geometric alignment with the source domain. Note that since possible keypoint configurations lie on a manifold with much lower dimension over the ambient space, the geometric alignment can provide effective regularization.

Our final formulation combines a standard supervised loss on the source domain with the two unsupervised regularization losses on view-consistency and geometric alignment. Our formulation can be easily optimized via alternating minimization and admits a simple strategy for variable initialization.

We evaluate the proposed approach on unsupervised domain adaptation from ModelNet~\cite{conf/cvpr/WuSKYZTX15} to rendered depth scans from the synthesized ShapeNet~\cite{DBLP:journals/corr/ChangFGHHLSSSSX15} 3D model dataset, and to real depth scans from the Redwood Object Scans~\cite{Choi2016} and 3DCNN Depth Scans~\cite{qi2016volumetric} datasets. Experimental results demonstrate that our approach can effectively reduce the domain gap between the online 3D model repositories and the real depth scans with background noise. Our approach is significantly better than without domain adaptation and is superior to general-purpose domain adaptation techniques such as ADDA~\cite{tzeng2017adversarial}. 
We also provide ablation studies to justify the design choice of each component of our approach.
Code is available at \url{https://github.com/xingyizhou/3DKeypoints-DA}.

\begin{figure*}[t]
\centering
\includegraphics[trim=0 220 0 220,clip,width=0.9\textwidth]{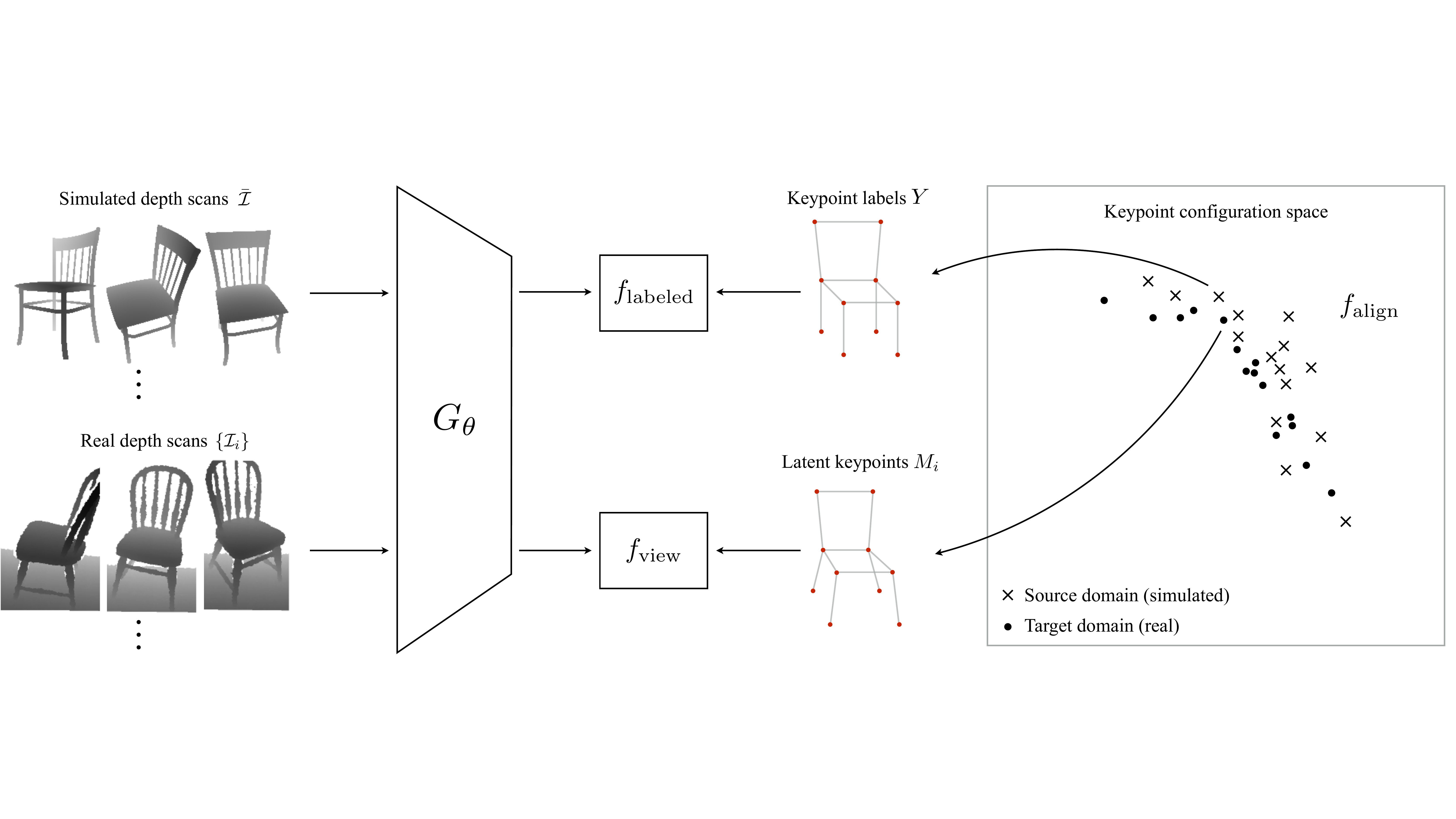}
\caption{\textbf{Approach overview.} We train an end-to-end 3D keypoint prediction network $G_{\theta}$ from labeled simulated depth scans $\bar{\mathcal{I}}$ of 3D models and unlabeled and unaligned real depth scans $\{\mathcal{I}_i\}$ of real world objects.}
\label{Figure:Approach:Overview}
\end{figure*}

\section{Related Works}
\label{Section:Related:Works}

\noindent\textbf{Keypoint Detection.} Keypoint detection from a single RGB or RGB-D image is a fundamental task in computer vision. We refer to~\cite{DBLP:conf/eccv/NewellYD16,Zhou_2017_ICCV,Simon_2017_CVPR,Tung_2017_ICCV} for some recent advances on this topic. While most techniques focus on developing novel neural network architectures for this task, fewer works focus on addressing the issue of domain shifts between the training data and testing data, e.g., the setting described in this paper. In ~\cite{Zhou_2017_ICCV}, the authors introduce a domain adaptation technique for 3D human pose estimation in the wild. Additionally for human pose estimation, \cite{Tung_2017_ICCV} proposes to align the source and target label distributions using a GAN loss. We opt to use an alternate metric that offers more flexibility in addressing domain shifts. Similarly to our method, \cite{Simon_2017_CVPR} also leverages the consistency across multiple views to boost the supervision on the target domain. However, the output of this approach is computed directly from the initial predictions from the source domain. In contrast, our approach only uses the initial predictions to initialize final predictions. Moreover, we utilize a latent configuration for synchronizing the predictions from multiple views, which avoids performing pair-wise analysis. 

\noindent\textbf{Multi-view supervision.} RGB and RGB-D video sequences essentially consist of different views of the same underlying 3D environment. In the literature, people have utilized such weak supervision for various tasks such as 3D reconstruction, novel view synthesis and 2D keypoint prediction, e.g.,\cite{DBLP:journals/corr/TulsianiZEM17,DBLP:journals/corr/YanYYGL16,DBLP:journals/corr/ZhaoWCLF17,DBLP:journals/corr/KalogerakisAMC16,Simon_2017_CVPR}. Our work differs from most works in the sense that we do not assume that relative poses between cameras are known. Instead, we introduce a pose invariant metric to compare keypoint configurations.
Concurrent to our work, Helge et al ~\cite{Rhodin_2018_CVPR} also introduced a similar viewpoint consistency term for un-supervised 3D human pose estimation. However, the multi-view data for articulated object is still hard to obtain. On the contrary, we use viewpoint consistency for rigid objects, where the views are free from videos. 

\noindent\textbf{Supervision from Big 3D Data.} Thanks the availability of annotated big 3D data such as ModelNet~\cite{conf/cvpr/WuSKYZTX15} and ShapeNet~\cite{DBLP:journals/corr/ChangFGHHLSSSSX15}, people have leveraged synthetic data generated from 3D models for various tasks, including image classification~\cite{conf/cvpr/WuSKYZTX15}, object recognition~\cite{DBLP:conf/eccv/SongX14,DBLP:conf/iccv/PengSAS15,DeepSlidingShapes}, semantic segmentation~\cite{zhang2016physically}, object reconstruction~\cite{song2016ssc,choy20163d,DBLP:conf/eccv/TatarchenkoDB16}, pose estimation~\cite{Su_2015_ICCV} and novel-view synthesis~\cite{DBLP:conf/iccv/SuWYG15,Zhou2016ViewSB}. The fundamental challenge of these approaches is that there are domain shifts between synthetic data and real RGB or RGB-D images. Most existing works focus on improving the simulation process to close this gap. In contrast, we focus on developing an unsupervised loss for domain adaptation. 

\noindent\textbf{Domain adaptation.} Domain adaptation~\cite{Busto_2017_ICCV,Gebru_2017_ICCV,Zhang_2017_ICCV,Gholami_2017_ICCV,Carlucci_2017_ICCV,Yan_2017_CVPR,Bousmalis_2017_CVPR,Herath_2017_CVPR,Koniusz_2017_CVPR,Tzeng_2017_CVPR,DBLP:journals/corr/Sankaranarayanan17a} for various visual recognition tasks is an active research area in computer vision, and our problem falls into the general category of domain adaptation. It is beyond the scope of this paper to provide a comprehensive review of the literature, however we refer to a recent survey ~\cite{DBLP:journals/corr/Csurka17} on this topic. A common strategy for unsupervised domain adaptation is to align the output distributions between source and target domains, e.g., either through explicit domain-wise maps or through use of a GAN. In contrast, our regularizations are tailored for the particular problem we consider, i.e., view-consistency and domain shifts caused by varying densities.

\section{Problem Statement}
\label{Section:Problem:Statement}

We study the problem of predicting complete 3D keypoints of an underlying object from a single image or depth scan.
We assume the input consists of a labeled dataset $\overline{\set{I}}$ and an unlabeled dataset $\set{I}$. Moreover, the unlabeled dataset is comprised of $N$ subsets $\set{I}_{i}, 1\leq i \leq N$, where each subset collects depth scans/images of the same object from different views. Such data naturally arises from RGB-D or RGB video sequences. 

Each instance $I\in \overline{\set{I}}$ in the labeled dataset possesses a ground-truth label $Y(I)\in \R^{3\times d}$, which is a matrix that collects the coordinates of the ordered keypoints in its columns. Without losing generality, we assume that the 3D local coordinate system of $I$ is chosen so that the centroid of $Y(I)$ is at the origin:
\begin{equation}
Y(I)\bs{1} = 0.
\label{Eq:Normalization}    
\end{equation}

It is expected that the source domain of the labeled dataset and the target domain of the unlabeled dataset are different (e.g., the source domain consists of synthetic images/scans but the target domain consists of real images/scans). Our goal is to train a neural network $G_{\theta}: \R^{m\times n}\rightarrow \R^{3 \times d}$ that takes an image from the target domain as input and outputs the predicted keypoints by leveraging both the labeled dataset $\overline{\set{I}}$ and unlabeled subsets $\set{I}_i, 1\leq i \leq N$. We define this problem as unsupervised domain adaptation for 3D keypoint prediction.

Note that we do not assume the underlying cameras of each unlabeled subset are calibrated, or in other words, the relative transformations between different views of the same object are not required. Although it is possible to align the depth scans to obtain relative transformations, we found that such alignments are not always reliable in the presence of scanning discontinuities where little overlaps between consecutive scans are available. In contrast, our formulation treats relative camera poses as latent variables, which are optimized together with the network parameters. 

\section{Approach}
\label{Section:Approach}

In this section, we describe our detailed approach to unsupervised domain adaptation for 3D keypoint prediction. We first introduce a pose-invariant distance metric to compare keypoint configurations in Section~\ref{Subsection:Pose:Invariant:Distance:Metric}. This allows us to compare the predictions in different views without knowing the relative transformations for uncalibrated datasets. We then present the formulation of our approach in Section~\ref{Subsection:Formulation}. Finally, we discuss our optimization strategy in Section~\ref{Subsection:Optimization}.

\subsection{Pose-Invariant Distance Metric}
\label{Subsection:Pose:Invariant:Distance:Metric}

The pose-invariant distance metric compares two keypoint configurations $X, Y \in \R^{3\times d}$ described in different coordinate systems. Since the mean of each keypoint configuration is zero, we introduce a latent rotation $R$ to account for the underlying relative transformation:
\begin{equation}
r(X,Y) = \min\limits_{R\in SO(3)} \|RX - Y\|_{\set{F}}^2,
\label{Eq:pose-inv-mrtric}
\end{equation}
where $\|\cdot\|_{\set{F}}$ denotes the matrix Frobenius Norm. It is clear that $r(X,Y)$ is independent of the coordinate systems associated with $X$ and $Y$, making it particularly suitable for comparing predictions from uncalibrated views and aligning the source domain and the target domain.

In the following, we discuss a few key properties of $r(X,Y)$ that will be used extensively in our approach. First of all, both $r(X,Y)$ and the gradient of $r(X,Y)$ with respect to each of its argument admit closed-form expressions. These are summarized in the following two propositions.
\begin{prop}  $r(X,Y)$ admits the following analytic expression:
$$
r(X,Y) = \|X\|_{\set{F}}^2 + \|Y\|_{\set{F}}^2 - 2\cdot \textup{trace}\big(R\cdot(XY^{T})\big) 
$$
where $R$ is derived from the singular value decomposition (or SVD) of $YX^{T} = U\Sigma V^{T}$:
\begin{equation}
R = U\diag(1,1,s)V^{T}, \quad s = \textup{sign}(\det(XY^{T})).
\label{Eq:R:OPT}
\end{equation}
\label{Prop:1}
\end{prop}
\noindent\textsl{Proof:} See~\cite{Horn87closed-formsolution}. \qed

\begin{prop} The gradient of $r(X,Y)$ with respect to $X$ is given by
$$
\frac{\partial r}{\partial X}(X,Y) = 2(X-R^{T}Y),
$$
where $R$ is given by Eq. (\ref{Eq:R:OPT}).
\label{Prop:2}
\end{prop}
\noindent\textsl{Proof:} Please refer to the supplemental material. \qed

Our optimization procedure also frequently involves the following optimization problem that computes the weighted average $X^{\star}$ of a set of keypoint configurations $Y_i, 1\leq i \leq n$ in the quotient space $\R^{3\times d}/SO(3)$:
\begin{align}
X^{\star}  = \underset{X\in \R^{3\times d}}{\textup{argmin}}\ 
\sum\limits_{i=1}^{n} c_i r(X, Y_i) = \underset{X\in \R^{3\times d}}{\textup{argmin}} \sum\limits_{i=1}^{n}c_i\min\limits_{R_i \in SO(3)}\|X-R_i^{T}Y_i\|_{\set{F}}^2,
\label{Eq:Quotient:Mean}
\end{align}
where $c_i, 1\leq i \leq n$ are constants. Although Eq.~(\ref{Eq:Quotient:Mean}) does not admit a closed-form solution, it can be easily optimized via alternating minimization. Specifically, when $X$ is fixed, each $R_i$ can be computed independently using Proposition~\ref{Prop:1}. When the $R_i$ latent variables are fixed, $X$ is simply given by the mean of $R_i^{T} Y_i$, i.e., $X = \frac{1}{\sum c_i}\sum\limits_{i=1}^{n}c_i R_i^{T} Y_i$. To make the solution unique, we always set $R_1 = I_3$. 

\subsection{Formulation}
\label{Subsection:Formulation}

\begin{figure}[t]
\begin{tabular}{ccc}
\includegraphics[width=0.32\textwidth]{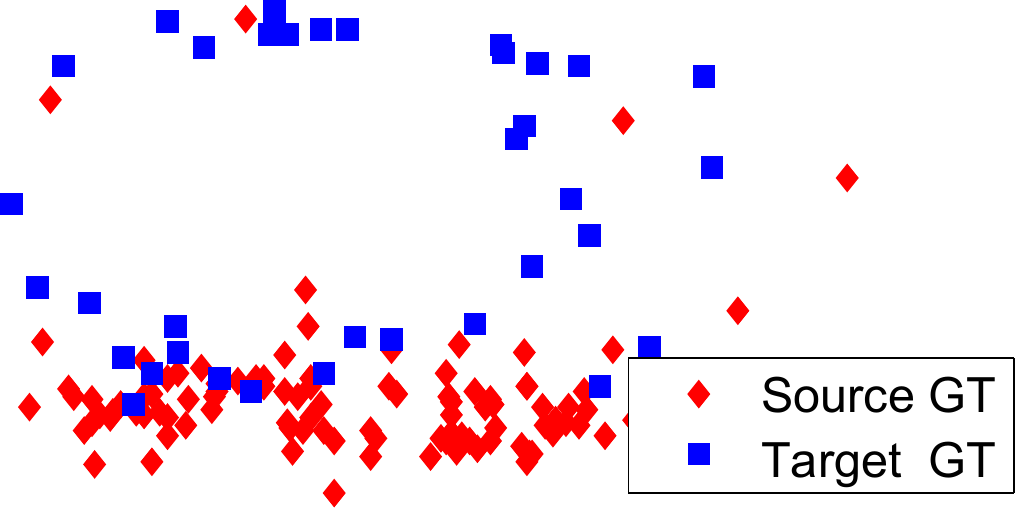}&
\includegraphics[width=0.32\textwidth]{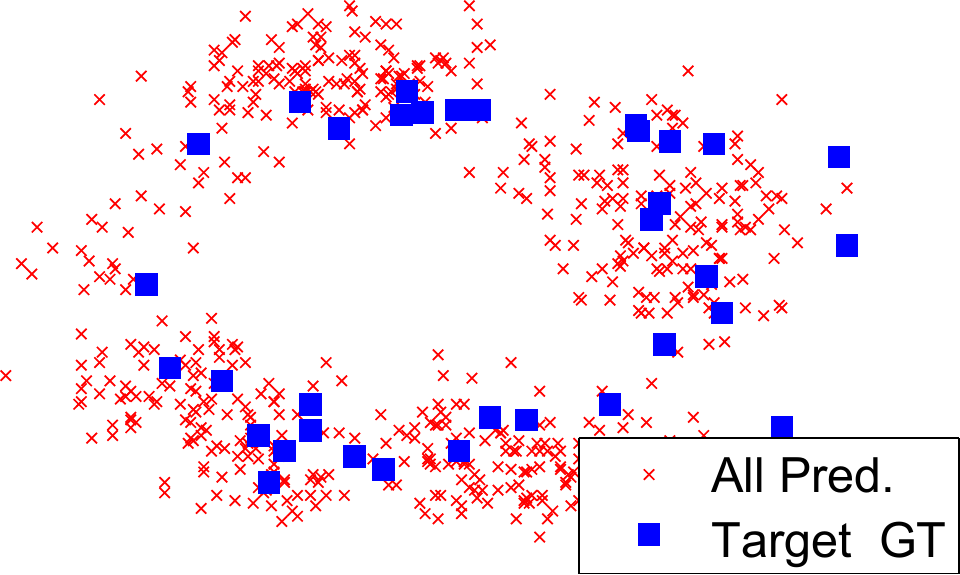}&
\includegraphics[width=0.32\textwidth]{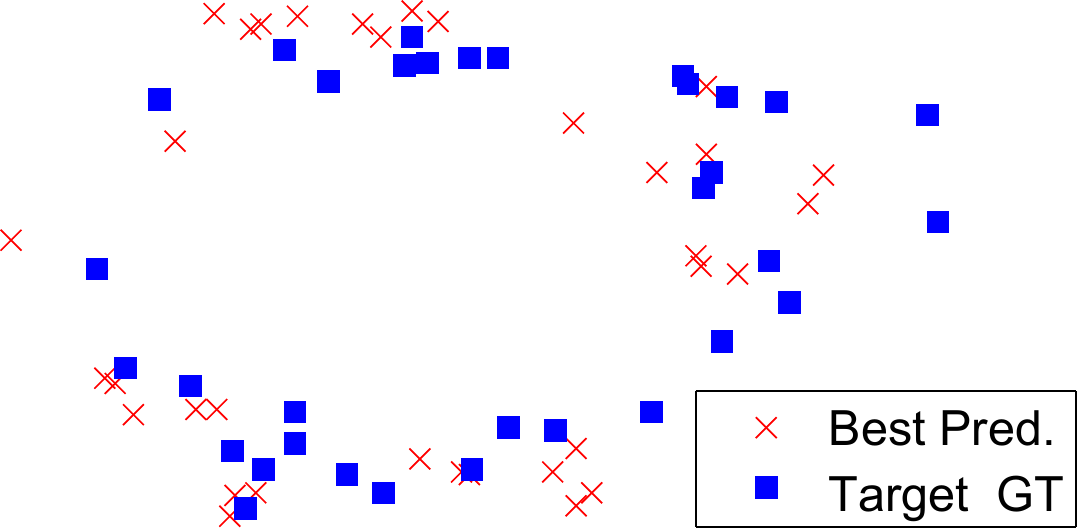}\\
(a) & (b) & (c)
\end{tabular}
\caption{\textbf{Latent Distribution and View Selection.} This figure provides visualizations of label distributions and view selection for initializing the latent configurations from ModelNet (source domain) to Redwood (target domain) on the Chair category. All visualizations are done by 2D projections using the first two principal components. \textbf{(a)} Label distributions of the source and target domains. 
\textbf{(b)} Visualizations of all predictions from different views. \textbf{(c)} Visualizations of the best prediction from each object. 
}
\label{Figure:Latent:Variable:Initialization}
\end{figure}

To train the keypoint prediction network $G_{\theta}(\cdot)$, we introduce three loss terms, namely, a labeled term $f_{\labeled}$, a view-consistency term $f_{\view}$ and a geometric alignment term $f_{\aligned}$.

The labeled term $f_{\labeled}$ fits predictions on the source domain labeled dataset $\overline{\set{I}}$ to the prescribed ground-truth labels. We use the regression loss under the L2-norm, which works well for 3D keypoint prediction tasks (c.f. ~\cite{Sun_2017_ICCV,Zhou_2017_ICCV}):
\begin{equation}
f_{\labeled} = \frac{1}{|\overline{\set{I}}|}\sum\limits_{I \in \overline{\set{I}}}\|G_{\theta}(I) - Y(I)\|_{\set{F}}^2.
\label{Eq:Labeled:Term}
\end{equation}

The view-consistency term $f_{\view}$ is defined on the target domain to enforce consistency between the predictions from different views of the same object. In other words, there exist pairwise rotations that transform the predictions from one view to another. A straightforward approach is to minimize $r(G_{\theta}(I_{ij}), G_{\theta}(I_{ij'}))$, where $I_{ij}$ and $I_{ij'}$ are different views of the same object. However, we found that such approach introduces a quadratic number of terms as the number of views increases and quickly becomes intractable. Therefore, we introduce a latent configuration $M_i \in \R^{3\times d}$ for each unlabeled subset $\set{I}_i$ that characterizes the underlying ground-truth in the canonical frame. We then define the view consistency term as:
\begin{equation}
f_{\view} = \frac{1}{N}\sum\limits_{i=1}^{N}\frac{1}{|\set{I}_i|}\sum\limits_{I_{ij} \in \set{I}_i}\ r(G_{\theta}(I_{ij}), M_i).
\label{Eq:Unlabeled:Term}
\end{equation}
It is clear that minimizing $f_{\view}$ automatically aligns the predictions across different views. The key advantages of Eq. (\ref{Eq:Unlabeled:Term}) over enforcing pairwise view-consistency are (i) the number of items is linear to the number of views, and (ii) as we will see immediately, the latent configurations $\{M_i\}$ allow us to easily formulate the geometric alignment term $f_{\aligned}$.

The geometric alignment term $f_{\aligned}$ prioritizes that the latent configurations $\{M_i,1\leq i\leq N\}$, which characterize the predictions on the target domain, shall be consistent with ground-truth labels $\{Y_{I}|I\in \overline{\set{I}}\}$ of the source domain. This term is conceptually similar to the idea of aligning output distributions for unsupervised domain adaptation, but our formulation is tailored to the specific problem we consider in this paper. A straightforward formulation is to use the Earth-Mover Distance between $\{M_i,1\leq i\leq N\}$ and $\{Y(I)|I\in \overline{\set{I}}\}$, which essentially aligns the two corresponding empirical distributions. However, we found that this strategy would force the alignment of keypoint configurations that are far apart, since the repetition counts of the same sub-types of an object may be different between the source and target domains (See Figure~\ref{Figure:Latent:Variable:Initialization}(a)). To address this issue, we propose to use the Chamfer distance for alignment:
\begin{equation}
f_{\aligned} = \frac{1}{N}\sum\limits_{i=1}^{N}\min\limits_{I\in \overline{\set{I}}}r(M_i, Y_{I}) + \frac{1}{|\overline{\set{I}}|}\sum\limits_{I \in \overline{\set{I}}}\min\limits_{1\leq i \leq N}\ r(M_i, Y_{I}).
\label{Eq:Align:Term}
\end{equation}
Intuitively, Eq.~(\ref{Eq:Align:Term}) still aligns the source and target domains, but it is insensitive to local density variations, and provides an effective way to address domain shifts.

We combine the labeled term $f_{\labeled}$, the view-consistency term $f_{\view}$ and the geometric alignment term $f_{\aligned}$ into the final loss function:
\begin{equation}
\underset{\theta,\{M_i\}}{\textup{minimize}} f_{\labeled} + \lambda f_{\view} + \mu f_{\aligned}.
\label{Eq:Total:Loss:Function}
\end{equation}
In our implementation, we set $\lambda = 1$ and $\mu = 0.1$.

\subsection{Optimization}
\label{Subsection:Optimization}

The major difficulty for optimizing Eq.~(\ref{Eq:Total:Loss:Function}) lies in the fact that the alignment term $f_{\aligned}$ is highly non-convex. In our experiments, we found that obtaining good initial values of the network parameters and latent variables is critical to achieving high-quality keypoint prediction network. In the following, we first introduce effective strategies to initialize the variables. We then show how to refine the variables using alternating minimization. 

\noindent\textbf{Network Parameter Initialization.} The network parameters are initialized by pre-training on the the source domain labeled dataset, i.e.,
\begin{equation}
\theta^{(0)} = \underset{\theta}{\min}\ \sum\limits_{I\in \overline{\set{I}}}\|G_{\theta}(I) - Y_{I}\|_{\set{F}}^2.
\end{equation}
It is then optimized via standard back-propagation.

\noindent\textbf{Latent Configuration Initialization.} We use the predictions obtained from the initial network $G_{\theta^{(0)}}(I_{ij}), I_{ij}\in \set{I}_i$ to initialize each latent variable $M_i$. To this end, we define a score for each prediction and set $M_i$ as the one with the highest score. The scoring function is motivated by the fact that the latent variables are expected to align with the source domain, we thus define an un-normalized density function:
\begin{equation}
p(M) = \sum\limits_{I \in \overline{\set{I}}}\exp(-\frac{r(M, Y(I))}{2\sigma^2}),
\label{Eq:Density:Definition}
\end{equation}
where $\sigma$ is chosen as mean of $r(G_{\theta^{(0)}}(I_{ij}), Y(I))$ between the predicted configurations and their closest labeled instances. Given Eq.~(\ref{Eq:Density:Definition}), we set
\begin{equation}
M_i^{(0)} = \underset{M\in \{G_{\theta^{(0)}}(I)|I\in \set{I}_i\}}{\textup{argmax}}\ p(M).
\label{Eq:M:init}
\end{equation}
As illustrated in Figure~\ref{Figure:Latent:Variable:Initialization}(b-c), this strategy leads to initial configurations that are close to the underlying ground-truth.

\noindent\textbf{Alternating Minimization.}
Given the initial network parameter $\theta^{(0)}$ and the initial latent configurations $M_i^{(0)}, 1\leq i \leq N$, we then refine them by solving Eq.~(\ref{Eq:Total:Loss:Function}) via alternating minimization. With $M_i^{(k)}$ and $\theta^{(k)}$ we denote their values at iteration k. At each alternating minimization step, we first fix the latent variables to optimize the network parameters. This leads to computing
\begin{align}
\theta^{(k+1)} = \underset{\theta}{\textup{argmin}}\  \frac{1}{|\overline{\set{I}}|}\sum\limits_{I\in \overline{\set{I}}}\|G_{\theta}(I)-Y_{I}\|_{\set{F}}^2 + \frac{\lambda}{N}\sum\limits_{i=1}^{N}\frac{1}{|\set{I}_i|}\sum\limits_{I\in \set{I}_i}r(G_{\theta}(I), M_i^{(k)}).      
\label{Eq:Altermin:Network:Opt}
\end{align}
Utilizing Proposition~\ref{Prop:2}, we apply stochastic gradient descent via back-propagation for solving Eq.~(\ref{Eq:Altermin:Network:Opt}).

We then fix the network parameters $\theta$ and optimize the latent variables $\{M_i^{(k+1)}\}$. In this case, Eq.~(\ref{Eq:Total:Loss:Function}) reduces to
\begin{align}
&\{M_i^{(k+1)}\} = \underset{\{M_i\}}{\textup{argmin}}\  \frac{\mu}{|\overline{\set{I}}|}\sum\limits_{I\in \overline{\set{I}}}\min\limits_{1\leq i \leq N}r(M_i, Y_{I}) \nonumber \\
&+ \frac{1}{N}\sum\limits_{i=1}^{N}\Big(\frac{\lambda}{|\set{I}_i|}\sum\limits_{I\in \set{I}_i}r(G_{\theta^{(k)}}(I),M_i)+\mu\min\limits_{I\in \overline{\set{I}}}r(M_i, Y_I)\Big).
\label{Eq:Altermin:Latent:Variable}
\end{align}
We again apply alternating minimization to solve Eq.~(\ref{Eq:Altermin:Latent:Variable}). In particular, we fix the closest point pairs given $\{M_i^{(k)}\}$ :
\begin{equation}
\hat{I}(i) = \underset{I\in \overline{\set{I}}}{\textup{argmin}} \ r(M_i^{(k)}, Y_I), \quad \hat{i}(I) =  \underset{1\leq i \leq N}{\textup{argmin}} \ r(M_i^{(k)}, Y_I).
\end{equation}
Given these closest pairs, we can optimize each latent configuration as
\begin{align}
\underset{M_i}{\textup{argmin}} \frac{\mu}{|\overline{\set{I}}|}\sum\limits_{I|\hat{i}(I)=i}r(M_i, Y_{I}) 
+ \frac{1}{N}\Big(\frac{\lambda}{|\set{I}_i|}\sum\limits_{I\in \set{I}_i}r(G_{\theta^{(k)}}(I),M_i)+\mu r(M_i, Y_{\hat{I}(i)})\Big) \label{Mi:OPT}.
\end{align}
Eq.~(\ref{Mi:OPT}) admits a form of Eq.~(\ref{Eq:Quotient:Mean}), and we apply the procedure described above to solve Eq.~(\ref{Mi:OPT}). 
In our experiments, we typically apply the inner alternating minimizations each 5 epochs for training the network parameters $\theta$.

\section{Evaluation}
\label{Section:Experiments}

For experimental evaluation, we first describe the experimental setup in Section~\ref{Subsection:Experimental:Setup}. We then present qualitative and quantitative results and compare our technique against baseline approaches in Section~\ref{Subsection:Analysis:Results}. We also present an ablation study to evaluate each component of our approach in Section~\ref{Subsection:Ablation:Study}. Finally, we further extend our method to 3D human pose estimation and RGB images in Section~\ref{Subsection:Experimental:Human} and Section~\ref{Subsection:Experimental:RGB}, respectively.

\subsection{Experimental Setup}
\label{Subsection:Experimental:Setup}

\noindent\textbf{Dataset.} Rendered depth scans of synthesized object models from the ModelNet~\cite{conf/cvpr/WuSKYZTX15} dataset serve as our source domain, and we test our domain adaptation method on three different target domains, namely: ShapeNet~\cite{DBLP:journals/corr/ChangFGHHLSSSSX15} (another synthesized 3D model dataset), the Redwood Object Scans real depth scan dataset~\cite{Choi2016}, and the 3DCNN real depth scan dataset~\cite{qi2016volumetric}. We focus our experiments on the chair, motorbike, and human classes, however we provide the most-detailed results on chairs because of their ubiquitousness across many popular 3D model and depth scan datasets. To provide keypoint labels for our source domain, we manually annotate the training samples in ModelNet with Meshlab~\cite{cignoni2008meshlab}.
To evaluate the accuracy of our system, we also annotate keypoints on our target domain datasets. 
This annotation is done by recovering each object's 3D mesh and each frame's camera pose from a depth video sequence. 
We only maintain frames in which all 2D projections of keypoints are within the image and keep the models with at least 20 valid frames. 
A summary of the four datasets used in our experiments is presented in Table~\ref{table:datasets}.
As a natural extension, we also test our method on the RGB images from the same Redwood dataset~\cite{Choi2016}.

\noindent\textbf{Data pre-processing.}
We assume the camera intrinsic and object's 3D bounding box are known both in training and testing depth images solely for data pre-processing.
We use the 2D projection of the 3D bounding box to crop each depth image. 
Additionally, the input depth images are centered by the mean depth and the depth values are normalized by the diagonal length of the 3D bounding box. 
Aside from the images, all keypoints are converted and evaluated in a unified coordinate system. 
Given a configuration, we subtract their mean and normalize by the diagonal length of the 3D bounding box.

\noindent\textbf{Evaluation protocol.} 
Similar to~\cite{DBLP:journals/corr/0001XLTTTF16}, we measure the Average distance Error (AE) between each predicted keypoint configuration and the corresponding annotation and plot the Percentage of Correct Keypoint (PCK) with respect to a threshold for each method for detailed comparison.
We also introduce a new metric, Pose-invariant Average distance Error (PAE) based on (\ref{Eq:pose-inv-mrtric}), for a better illustration of how our proposed method works.
The AE and PAE are shown in percentage and represent the relative ratio to the diagonal length of the 3D bounding box.

\noindent\textbf{Baseline methods.} We consider three baseline methods for experimental evaluation. 
\begin{itemize}
\item \textbf{Baseline I.} We first test  performance without any domain adaptation techniques, namely we directly apply the keypoint predictor trained on the source domain to the target domain. This baseline serves as a performance lower bound for accessing domain adaptation techniques.

\item \textbf{Baseline II.} We implement a state-of-the-art deep unsupervised general domain adaptation technique described in~\cite{tzeng2017adversarial}, 
which encourages domain confusion by fine-tuning the feature extractor on the target domain.

\item \textbf{Baseline III.} We apply supervised keypoint prediction on the target domain. To this end, we annotate 50 additional models from each domain and fine-tune Baseline I on these labeled instances. This baseline serves as a performance upper bound for accessing domain adaptation techniques. 
\end{itemize}
In Table~\ref{table:baselines} we compare these baselines to our approach on the Chair dataset. In addition, we provide before/after adaptation results for motorbike and human in Table~\ref{table:more}. We also conduct an ablation study on the Chair dataset to evaluate each component of our approach (Table~\ref{table:ablation} and Figure~\ref{Figure:Ablation:Study}).

\noindent\textbf{Implementation details.} We use ResNet50~\cite{he2016deep} pre-trained on ImageNet as our keypoint prediction network $G_\theta$. 
In order to fit our depth scans to the ResNet50 input (and additionally, to allow for natural extension to the RGB image domain), we duplicate the depth channel three times.
The network is first trained on source domain  $\overline{\set{I}}$ for 120 epochs, and then fine-tuned on a specific target domain $\set{I}$ for 30 epochs. 
The network is trained using a SGD optimizer via back-propagation, with learning rate 0.01 (dropped to 0.001 after 20 epochs), batch size 64, momentum 0.9 and weight decay 1e-4, which are all the default parameters in Resnet50~\cite{he2016deep}.
Our implementation is done in PyTorch.

\subsection{Analysis of Results}
\label{Subsection:Analysis:Results}

Table~\ref{table:baselines}, Table~\ref{table:more}, Table~\ref{table:ablation}, Figure~\ref{Figure:Ablation:Study}, and Figure~\ref{Figure:Results} present the quantitative and qualitative results of our approach.

\setlength{\tabcolsep}{2pt}
\begin{table*}
\begin{center}
\caption{\textbf{Results of our proposed methods tested on chairs after domain adaptation on different target domains.} We show Average distance Error (AE) and Pose-Invariant Average distance Error (PAE) in percentage. For both metrics, the lower the better.}
\resizebox{.95\columnwidth}{!}{%
\begin{tabular}{lcccccccc}
\noalign{\smallskip}
\hline
\noalign{\smallskip}
Target-Metric & Default-AE & ADDA-AE & Ours-AE & Supervised-AE & Default-PAE & ADDA-PAE & Ours-PAE & Supervised-PAE\\
\noalign{\smallskip}
\hline
\noalign{\smallskip}
ModelNet~\cite{conf/cvpr/WuSKYZTX15} & - & - & - & 5.56 & - & - & - & 4.76\\
ShapeNet~\cite{DBLP:journals/corr/ChangFGHHLSSSSX15} & 6.97 & 6.98 & \textbf{6.60} & 5.82 & 5.77 & 5.89 & \textbf{5.32} & 4.77\\
RedwoodDepth~\cite{Choi2016} & 16.01 & 15.44 & \textbf{12.76} & 8.67 & 10.73 & 10.13 & \textbf{8.27} & 5.68\\
3DCNN~\cite{qi2016volumetric} & 11.61 & 11.81 & \textbf{10.60} & 6.73 & 8.15 & 8.19 & \textbf{7.25} & 4.98\\
\noalign{\smallskip}
\hline
\noalign{\smallskip}
RedwoodRGB~\cite{Choi2016} & 27.59 & 26.16 & \textbf{25.24} & 11.90 & 13.44  & 12.31 & \textbf{11.38} & 7.67\\
\hline
\end{tabular}
}
\label{table:baselines}
\end{center}
\end{table*}
\setlength{\tabcolsep}{1.4pt}

\begin{table}
\begin{minipage}[b]{0.48\linewidth}
    \centering
    \caption{\textbf{Quantitative Results - AE}}
    \resizebox{.9\columnwidth}{!}{%
    \begin{tabular}{lcc}
    \hline\noalign{\smallskip}
    Category & Motorcycle & Human \\
    \noalign{\smallskip}
    \hline
    \noalign{\smallskip}
    Before adaptation & 21.55\% & 153.39mm \\
    After adaptation & 18.92\% & 135.56mm\\ 
    Supervised & 16.17\% & 113.44mm\\ 
    \hline
    \end{tabular}
    }
    \label{table:more}
\end{minipage}\hfill
\begin{minipage}[b]{0.48\linewidth}
    \caption{\textbf{Statistics of the datasets}.}
    \centering
    \resizebox{.9\columnwidth}{!}{%
    \begin{tabular}{lcccccccc}
    \hline\noalign{\smallskip}
    Target & \#Train Models & \#Test Models & Avg \#frames \\
    \noalign{\smallskip}
    \hline
    \noalign{\smallskip}
    ModelNet\cite{conf/cvpr/WuSKYZTX15} & 899 & 100 & Inf \\ 
    ShapeNet~\cite{DBLP:journals/corr/ChangFGHHLSSSSX15} & 2500 & 100 & Inf \\ 
    Redwood~\cite{Choi2016} & 200 & 35 & 150 \\ 
    3DCNN~\cite{qi2016volumetric} & 9 & 3 & 80 \\ 
    \hline
    \end{tabular}
    }
    \label{table:datasets}
\end{minipage}
\end{table}

\noindent\textbf{Qualitative results.} As illustrated in Figure~\ref{Figure:Results}, our approach yields keypoint structures that are consistent with the underlying ground-truths. Even under significant background noise and incomplete observations, our approach leads to faithful structures. 
Exceptions include the case for chair types that involve swivel bases. 
In this case, the predicted legs may be tilted. 
This is expected since the annotations may become unreliable in cases when the legs do not fall directly below the seat corners.

\noindent\textbf{Quantitative assessment.} 
As shown in Table~\ref{table:baselines}, the mean deviations of our approach in the two real depth scan datasets Redwood~\cite{Choi2016} and 3DCNN~\cite{qi2016volumetric} for the chair object class are 12.76\% and 10.60\% of the diagonal length of object bounding box, respectively. 
This translates to approximately 7-10 cm, which is fairly accurate when compared to the radius of a chair's base. Additional experiments done on the motorbike class yield similar improvements, as indicated by Table~\ref{table:more}. For the motorbike training process, we utilize the ShapeNet dataset as our source domain and the Redwood dataset as our target domain.

\noindent\textbf{Analyses of performance across different datasets.}
Table~\ref{table:baselines} shows that our method gives consistent performance improvements on all three target depth domains.
For the synthesized dataset ShapeNet~\cite{DBLP:journals/corr/ChangFGHHLSSSSX15}, which has a relatively small domain shift from the supervised training set, our unsupervised terms are still able to push error rates close to the supervised upper bound.
The advantages of our proposed method can be best observed in the Redwood dataset~\cite{Choi2016}, where using our full error terms leads to a 44\% step towards the supervised performance upper-bound.
Additionally, the improvement in 3DCNN Dataset~\cite{qi2016volumetric} is still decent despite the very limited available models and poor depth image quality.

\noindent\textbf{Analysis of performance gain.}
Our performance gains can be attributed to our network learning more plausible keypoint configuration shapes, which is supported by the fact that the improvement of AE is always close to that of PAE. This is expected because our unsupervised terms are viewpoint-invariant and focus on improving the keypoint configuration shape.

\noindent\textbf{Comparison to ADDA~\cite{tzeng2017adversarial}.} 
Our approach is superior to the state-of-the-art unsupervised domain adaptation technique~\cite{tzeng2017adversarial} in the keypoint estimation task. ADDA aims to cross the domain gap by aligning the feature distributions of the source and target domains, which is complementary to our approach's constraints on the label space.
We argue that there is more structure to rely on in label space than feature space for rigid objects.
Another important factor is that view consistency is not incorporated in ADDA~\cite{tzeng2017adversarial}.

\subsection{Ablation Study}
\label{Subsection:Ablation:Study}

We present ablation studies to justify each component of our approach. We restrict our study to a sole object class, chair, and to the representative target domains, ShapeNet and Redwood Object Scans.

\begin{table}
\caption{\textbf{Chair ablation study on ShapeNet and Redwood Object Scans dataset.} We show the Average distance Error (AE) in percentage for each approach, including the three baselines.}
\begin{minipage}{0.5\linewidth}
    \centering
    \includegraphics[width=0.8\textwidth]{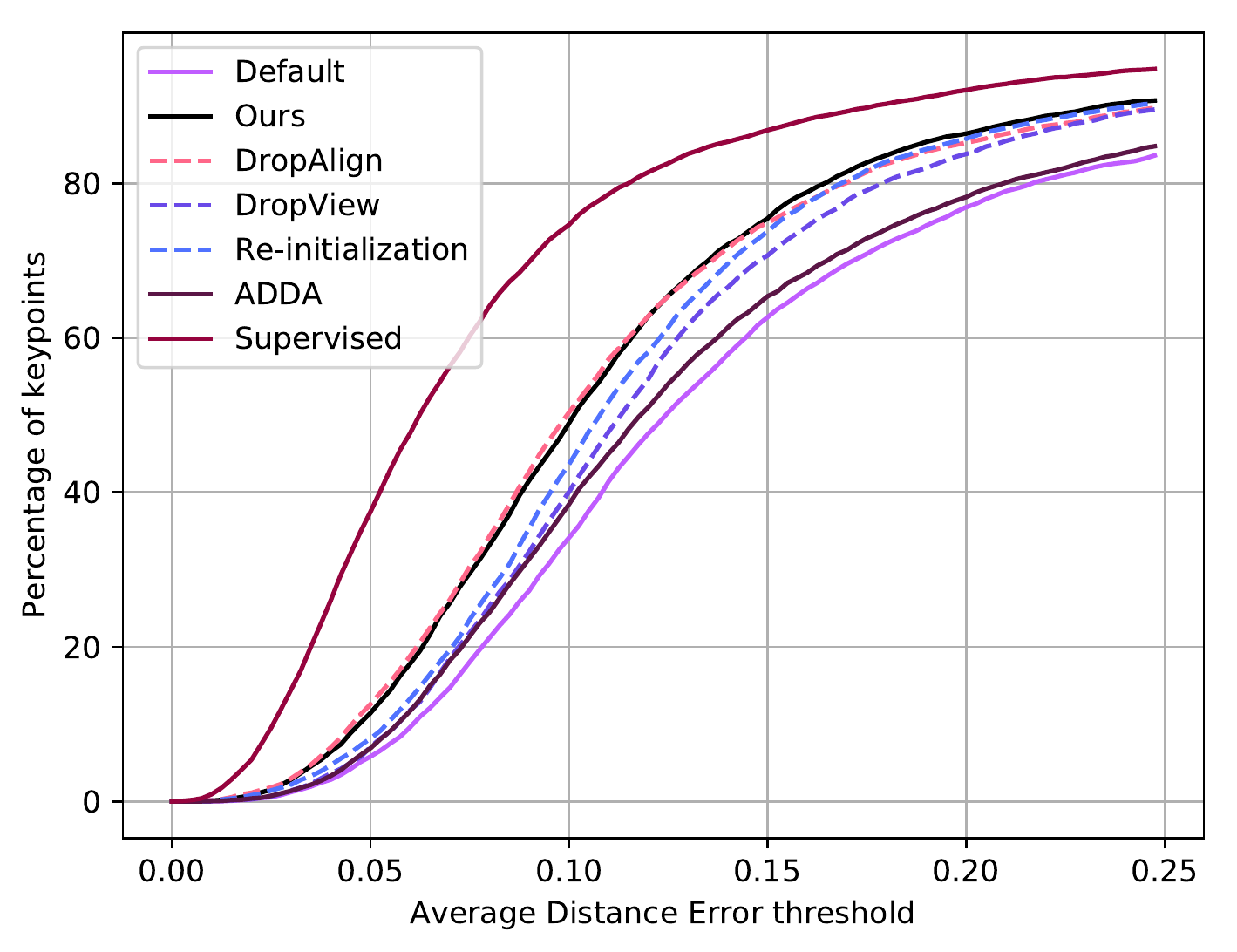}
    \label{Figure:Ablation:Study}
\end{minipage}\hfill
\begin{minipage}{0.5\linewidth}
    \centering
    
    \resizebox{\columnwidth}{!}{%
    \begin{tabular}{lcccccccc}
    \hline\noalign{\smallskip}
    Target domain & ShapeNet(\%)  & Redwood depth(\%)\\
    \noalign{\smallskip}
    \hline
    \noalign{\smallskip}
    Ours & \textbf{6.60} & \textbf{12.76} \\
    Drop view & 6.70 & 13.95 \\
    Drop align & 6.67 & 12.97 \\
    Re-initialize & 6.66 & 13.43 \\
    \noalign{\smallskip}
    \hline
    \noalign{\smallskip}
    Default & 6.97 & 16.01 \\
    ADDA~\cite{tzeng2017adversarial} & 6.98 & 15.44 \\
    Supervised lower bound & 5.82 & 8.67 \\
    \hline
    \end{tabular}
    }
    \label{table:ablation}
\end{minipage}
\captionof{figure}{\textbf{Baseline \& Ablation Study.} Comparisons between our approach with alternative approaches on Redwood depth Dataset~\cite{Choi2016}. The Figure shows Percentage of Correct Keypoints (PCK) under a threshold.}
\end{table}

\noindent\textbf{Dropping the view-consistency term.} 
We test the effects of dropping the view-consistency term. 
In this case, we simply align the output from all the depth scans with annotations of the source domain. 
As shown in Table~\ref{table:ablation} and Figure~\ref{Figure:Ablation:Study}, the performance drops considerably compared to our full term, while still maintaining better performance than without adaptation. Thus, if the predictions on the majority of views are consistent with one another, the keypoint configuration obtained by averaging all the predictions can serve as a reliable guidance to correct the bad outliers.

\noindent\textbf{Dropping the alignment term.} 
Without output alignment, merely utilizing the view consistency term can also significantly reduce the testing error.
This can be interpreted as the network updating the latent variables in a self-guided manner, based solely on the consistency between different views.

\noindent\textbf{Latent configuration updates versus re-initialization}
Instead of updating the latent configurations $M_i$ by solving Eq.~\ref{Mi:OPT}, we can apply Eq.~\ref{Eq:M:init} to re-initialize the latent configurations, which is also consistent with our training framework. 
The results is worse than updating $M_i$ by minimizing the view-consistency term, 
showing an advantage of our alternating minimization schema.

\subsection{Extension to Human Pose}
\label{Subsection:Experimental:Human}
\begin{figure}[t]
\centering
\setlength{\tabcolsep}{0pt}
\resizebox{.85\columnwidth}{!}{%
\begin{tabular}{|cc|cc|cc|cc|cc|}
\hline
\includegraphics[trim=20 0 20 0,clip,width=0.24\linewidth]{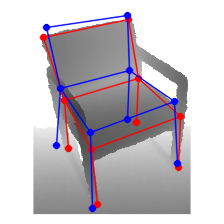} &
\includegraphics[trim=20 0 20 0,clip,width=0.24\linewidth]{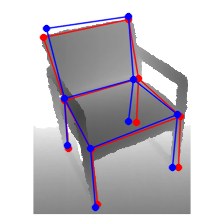} &
\includegraphics[trim=20 0 20 0,clip,width=0.24\linewidth]{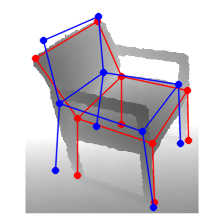} &
\includegraphics[trim=20 0 20 0,clip,width=0.24\linewidth]{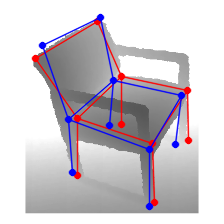} &
\includegraphics[trim=20 0 20 0,clip,width=0.24\linewidth]{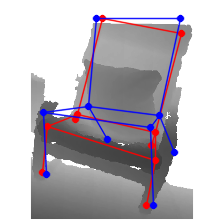} &
\includegraphics[trim=20 0 20 0,clip,width=0.24\linewidth]{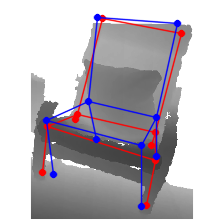} &
\includegraphics[trim=20 0 20 0,clip,width=0.24\linewidth]{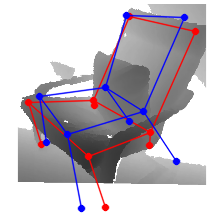} &
\includegraphics[trim=20 0 20 0,clip,width=0.24\linewidth]{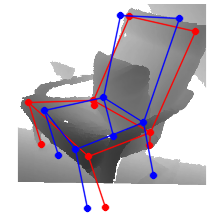}\\ [-4pt]
\hline
\includegraphics[trim=20 0 20 0,clip,width=0.24\linewidth]{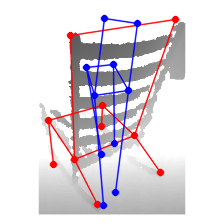} &
\includegraphics[trim=20 0 20 0,clip,width=0.24\linewidth]{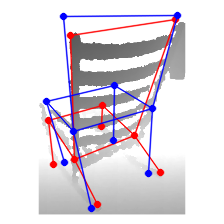} &
\includegraphics[trim=20 0 20 0,clip,width=0.24\linewidth]{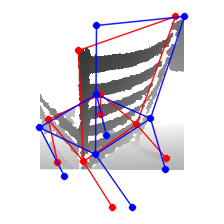} &
\includegraphics[trim=20 0 20 0,clip,width=0.24\linewidth]{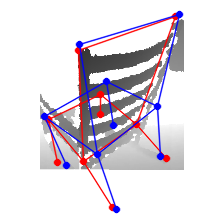} &
\includegraphics[trim=20 0 20 0,clip,width=0.24\linewidth]{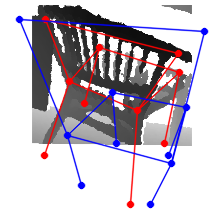} &
\includegraphics[trim=20 0 20 0,clip,width=0.24\linewidth]{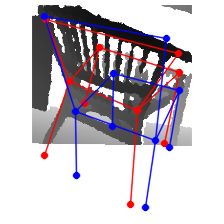} &
\includegraphics[trim=20 0 20 0,clip,width=0.24\linewidth]{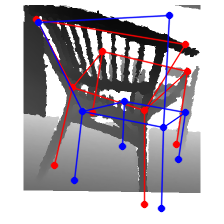} &
\includegraphics[trim=20 0 20 0,clip,width=0.24\linewidth]{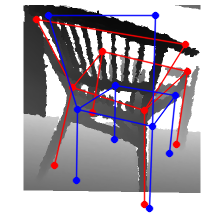}\\ [-4pt]
\hline
\includegraphics[trim=20 0 20 0,clip,width=0.24\linewidth]{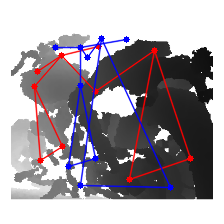} &
\includegraphics[trim=20 0 20 0,clip,width=0.24\linewidth]{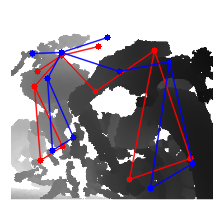} &
\includegraphics[trim=20 0 20 0,clip,width=0.24\linewidth]{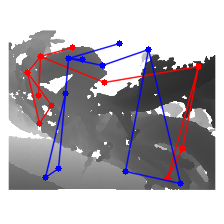} &
\includegraphics[trim=20 0 20 0,clip,width=0.24\linewidth]{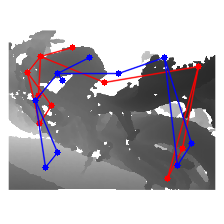} &
\includegraphics[trim=20 0 20 0,clip,width=0.24\linewidth]{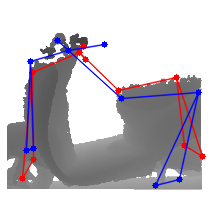} &
\includegraphics[trim=20 0 20 0,clip,width=0.24\linewidth]{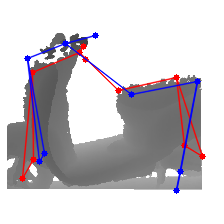} &
\includegraphics[trim=20 0 20 0,clip,width=0.24\linewidth]{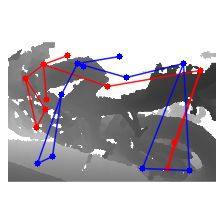} &
\includegraphics[trim=20 0 20 0,clip,width=0.24\linewidth]{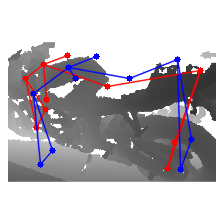}\\ [-4pt]
\hline
\includegraphics[trim=20 0 20 0,clip,width=0.24\linewidth]{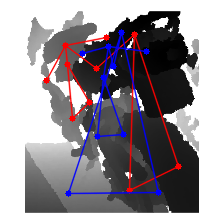} &
\includegraphics[trim=20 0 20 0,clip,width=0.24\linewidth]{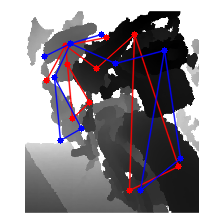} &
\includegraphics[trim=20 0 20 0,clip,width=0.24\linewidth]{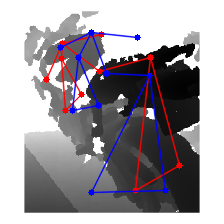} &
\includegraphics[trim=20 0 20 0,clip,width=0.24\linewidth]{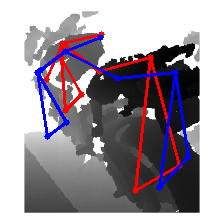} &
\includegraphics[trim=20 0 20 0,clip,width=0.24\linewidth]{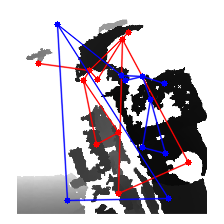} &
\includegraphics[trim=20 0 20 0,clip,width=0.24\linewidth]{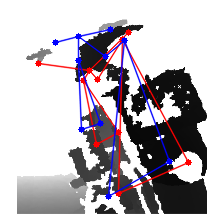} &
\includegraphics[trim=20 0 20 0,clip,width=0.24\linewidth]{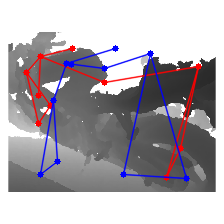} &
\includegraphics[trim=20 0 20 0,clip,width=0.24\linewidth]{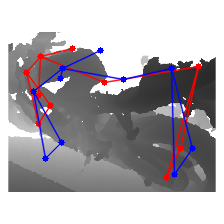} \\ [-4pt]
\hline
\end{tabular}
}
\\
\resizebox{.85\columnwidth}{!}{%
\begin{tabular}{|cc|cc|cc|cc|}
\hline
\includegraphics[trim=20 0 20 0,clip,width=0.24\linewidth]{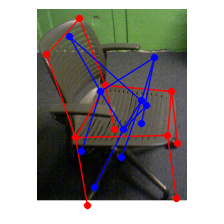} &
\includegraphics[trim=20 0 20 0,clip,width=0.24\linewidth]{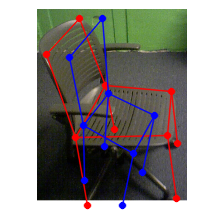} & 
\includegraphics[trim=20 0 20 0,clip,width=0.24\linewidth]{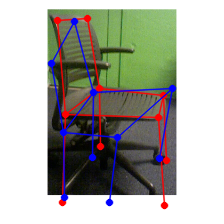} &
\includegraphics[trim=20 0 20 0,clip,width=0.24\linewidth]{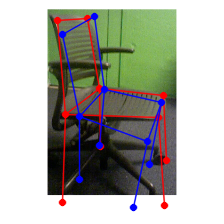}
\includegraphics[trim=20 0 20 0,clip,width=0.24\linewidth]{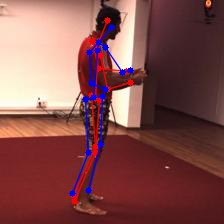} &
\includegraphics[trim=20 0 20 0,clip,width=0.24\linewidth]{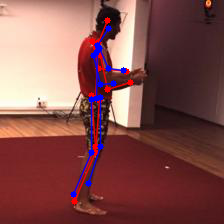} &
\includegraphics[trim=20 0 20 0,clip,width=0.24\linewidth]{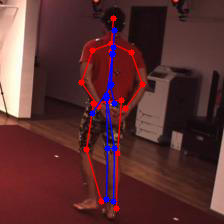} &
\includegraphics[trim=20 0 20 0,clip,width=0.24\linewidth]{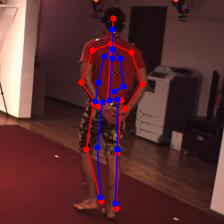}\\ [-4pt]
\hline
\end{tabular}
}
\caption{\textbf{Qualitative results.} We compare 3D keypoint predictions (blue) before (left) and after (right) using our approach on different datasets. For each model we show 2 views. Reference ground-truth are in red. 
}
\label{Figure:Results}
\end{figure}

Additionally, we perform experiments for human keypoints using the Human 3.6M dataset~\cite{h36m_pami}. The Human 3.6M dataset~\cite{h36m_pami} provides 3D human joint annotations for 7 subjects (5 for training and 2 for testing) from 4 different camera views. We use 3 of the 5 training subjects as supervised (source) samples and the remaining 2 training subjects as unsupervised (target) samples, trained with the proposed multi-view consistency and output alignment constraints. The result is shown in Table~\ref{table:more} and Figure~\ref{Figure:Results}. The supervised performance upper-bound of our implementation is $113.44mm$, which approximately matches the \emph{3D data-only} state-of-the-art~\cite{Sun_2017_ICCV}.

\subsection{Extension to RGB images}
\label{Subsection:Experimental:RGB}
Our approach can seamlessly be applied to keypoint estimation from RGB images.
We show our preliminary results on Table~\ref{table:baselines}, which indicate that our proposed method is able to reduce the AE from the baseline without domain adaptation. 
As shown in Figure~\ref{Figure:Results}, our method helps regularize the output when the before-adaptation baseline predicts a seemingly random point set.
\section{Conclusions}
\label{Section:Conclusions}
In this paper, we introduced an unsupervised domain adaptation approach for keypoint prediction from a single depth image. Our approach combines two task-specific regularizations, i.e., view-consistency and label distributions alignment of the source and target domains. Experimental results show that our approach is significantly better than without domain adaptation and is superior to state-of-the-art generic domain adaptation methods. Additionally, our multi-view consistency and output alignment terms makes it easier to leverage mass amounts of unlabeled 3D data for 3D tasks such as viewpoint estimation and object reconstruction.

\noindent\textbf{Acknowledgement.} Qixing Huang would like to acknowledge support of this research from NSF DMS-1700234, a Gift from Snap Research, and a hardware Donation from NVIDIA. 

\clearpage

\bibliographystyle{splncs04}
\bibliography{keypoint}
\end{document}